\newif\ifblind
\newif\ifpreprint

\blindfalse \preprinttrue

\documentclass{article}
\usepackage{arxiv}

\usepackage[utf8]{inputenc} \usepackage[T1]{fontenc}    \usepackage{hyperref}       \usepackage{url}            \usepackage{booktabs}       \usepackage{amsmath}
\usepackage{amsfonts}       \usepackage{nicefrac}       \usepackage[nopatch=footnote]{microtype}      \usepackage{cleveref}       \usepackage{acronym}

\newacro{IDM}{item-difficulty modelling}
\newacro{IRT}{item response theory}
\newacro{LLTM}{linear logistic test model}
\newacro{MCQ}{multiple-choice question}
\newacro{MCQA}{multiple-choice question-answering}
\newacro{MTL}{multi-task learning}
\newacro{LLM}{large language model}
\newacro{BERT}{Bidirectional Encoder Representations from Transformers}
\newacro{RMSE}{root mean square error}
\newacro{NLP}{natural language processing}
\newacro{MLM}{masked language modelling}
\newacro{MSE}{mean-squared error}
\newacro{CI}{confidence interval}
\newacro{API}{Application Programming Interface}
\newacro{GELU}{Gaussian Error Linear Unit}
\newacro{TFIDF}[TF-IDF]{term frequency-inverse document frequency}

\newcommand{\rev}[1]{#1}

\usepackage{graphicx}
\usepackage{tikz}
\usetikzlibrary{positioning, arrows.meta, fit, calc, shapes.geometric}
\usepackage{subcaption}
\usepackage{setspace}
\usepackage{csquotes}
\usepackage[
  backend=biber,
  style=apa,
  natbib
]{biblatex}
\title{Response-free item difficulty modelling for multiple-choice items with fine-tuned transformers: Component-wise representation and multi-task learning}
\renewcommand{\shorttitle}{Transformer-based IDM} 

\ifblind
  \author{}
\else

\usepackage{authblk}

\newbox{\orcid}\sbox{\orcid}{\includegraphics[scale=0.06]{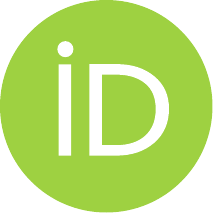}}
\author[1,2]{%
  \href{https://orcid.org/0000-0002-3888-3203}{\usebox{\orcid}\hspace{1mm}Jan Netík}%
}
\author[1,2]{%
  \href{https://orcid.org/0000-0003-4754-8543}{\usebox{\orcid}\hspace{1mm}Patrícia Martinková}%
}
\affil[1]{Faculty of Education, Charles University, Prague, Czech Republic}
\affil[2]{Institute of Computer Science of the Czech Academy of Sciences, Prague, Czech Republic}

\hypersetup{pdfauthor={Jan Netík, Patrícia Martinková}}

\fi

\ifpreprint\else
    \renewcommand{\headeright}{}
    \renewcommand{\undertitle}{}
  \fi

\hypersetup{
  pdftitle={Response-free item difficulty modelling for multiple-choice items with fine-tuned transformers: Component-wise representation and multi-task learning},
  pdfkeywords={item difficulty modelling, transformers, fine-tuning, multiple-choice items, reading comprehension, Rasch model},
}

\begin{document}

\singlespacing

\maketitle

\begin{abstract}

\rev{
Item difficulty must often be estimated before test administration, when no responses are yet available for calibration. While most response-free difficulty modelling approaches derive item-text features by hand for a separate statistical model, we fine-tune a transformer end-to-end on the wording, avoiding the theory-based feature design and the preprocessing that discards information. We address reading-comprehension multiple-choice items, whose difficulty depends on inferential demands spanning passage, question, and options, yet the simplest model sees one undifferentiated sequence and is trained on difficulty alone. We introduce and investigate two extensions to the joint-encoding baseline: a component-wise variant, which encodes the wording parts separately, and a multi-task variant, which adds an auxiliary task of question answering. We compare the methods across three training-set sizes sampled from a corpus of nearly 30,000 items whose labels approximate response-based Rasch difficulty. At the smallest training size, both extensions improve on the baseline, the multi-task variant across every metric, and component-wise encoding in rank ordering. Further research may ground the auxiliary supervision in observed responses and extend the approach to other item types.
}
\end{abstract}

\noindent\textbf{Keywords:} item difficulty modelling; transformers; fine-tuning; multiple-choice items; reading comprehension; Rasch model

\acresetall

\section{Introduction}

Accurate estimation of item difficulty is \rev{fundamental to} psychometric modelling, particularly in high-stakes educational assessment. Reliable difficulty estimates are essential for minimising measurement error at targeted ability levels, aligning the difficulty of multiple test forms during construction, and providing test-takers with optimally selected items in computerized adaptive testing \rev{\autocite{martinkova2023computational}}. In many practical settings, however, conventional estimation from a sufficient number of response patterns is unavailable: newly written items may not yet have been administered, pre-testing can expose secure items, the target population may be too small or difficult to recruit, and automatically generated item pools may need provisional difficulty estimates before response data are available. \rev{This last case is the operational setting for automatic item generation, where the system can yield far more items than can realistically be piloted, so that difficulty has to be pre-estimated from the generated wording itself \autocite{gierl2013, attali2022}.} Expert judgments can partly fill this gap, but they are not sufficiently reliable as a sole source of information \autocite{alkhuzaey2023, stepanek2023}. This creates a need for methods that predict item difficulty directly from the item stimulus material, that is, from the wording available before administration.

We refer to this pre-administration setting as \emph{response-free} \ac{IDM}: item difficulty is inferred from the item's wording without requiring response patterns \citep{ulitzsch2025}. Three lines of work approach the task \autocite[see][for recent systematic reviews]{peters2025, alkhuzaey2023}\rev{; for a review oriented specifically to content-informed \ac{IDM} with machine-learning and \ac{LLM} methods, see \autocite{MartinkovaPotuznikovaNetik2026_chapter}}. The oldest manually defines and derives so-called \textit{item text features} \autocite[readability indices, syntactic depth, lexical complexity, \ac{TFIDF} weightings of word counts, or, as in][, averaged static word embeddings]{stepanek2023} and feeds them to a statistical or machine-learning model, typically after demanding preprocessing \autocite[stopword removal, lemmatisation;][]{stepanek2023, freedle1993} that discards a substantial morphosyntactic layer of the wording. The pipeline is highly interpretable but replaces the wording's raw content with a hand-engineered feature representation, requiring substantial preprocessing and theory-based feature-design effort. The second line, rooted in explanatory \acl{IRT}\acused{IRT} \rev{\autocite[\ac{IRT};][]{deboeck2004}}, integrates fixed item features directly into the \ac{IRT} likelihood: in the linear logistic test model \citep[LLTM;][]{fischer1973}\acused{LLTM} the item text features are hand-coded as in the first approach and their ``weights'' are estimated, while also leveraging the response-patterns simultaneously. Some works extended the same framework using \emph{frozen} contextual representations from a pre-trained transformer \autocite[specifically, the encoder-only version called ``\acl{BERT}'', \acs{BERT}\acused{BERT}; for the original architecture see][]{vaswani2017} as feature vectors for \textit{cloze}-style\footnote{\rev{The \textit{cloze} task, introduced by \citet{taylor1953a} is, in effect, identical to the \ac{MLM} objective on which \ac{BERT}-family encoders are pre-trained \autocite{devlin2019}: a token is hidden in context and the system must recover it, exactly as a test-taker would.}} items \autocite{mccarthy2021, yancey2024}; the \ac{LLTM} specification is retained, but the hand-coded features are replaced by transformer outputs. A third line fine-tune\rev{s the encoder} end-to-end against the difficulty target, so \rev{that its} representations are adapted \rev{specifically} to the \ac{IDM} task rather than borrowed from a general-purpose pre-training objective. \rev{The choice among these three approaches depends on the assessment domain, item format, available training sample, and whether interpretability or predictive accuracy is the primary objective \autocite{MartinkovaPotuznikovaNetik2026_chapter}.}

\rev{In this study, we pursue the third, end-to-end approach.} The fine-tuning \rev{procedure} is well established in the broader \ac{NLP} \rev{literature} \autocite{howard2018finetuning, devlin2019}, and has \rev{previously} been applied to \ac{MCQ} \ac{IDM} \autocite{netik2024}. \citet{benedetto2021} reported an independent adaptation in a different setting (short factoid-style items: computer-science trivia and K-12 mathematics problems, with limited passage context). The line has not been substantially extended for reading-comprehension \ac{MCQ} since. \rev{While item difficulty dominates the text-based prediction literature and is the sole item parameter considered in the systematic reviews we examined, other parameters are also important for understanding item functioning. Discrimination indicates how effectively an item differentiates among respondents at different ability levels, while lower and upper asymptote parameters can capture response processes associated with guessing and inattention, respectively \autocite{pavlech2026bridging4P}. These parameters have received considerably less attention in content-informed modelling. One relevant study is \textcite{benedetto2020}, who estimated item discrimination alongside difficulty from question and option text using \ac{TFIDF} features and random-forest regressors. Predicting these additional parameters may be especially demanding because they reflect how response probabilities vary across the ability continuum, rather than an aggregate response tendency across respondents, and are generally estimated less precisely than item difficulty.}

Response-free \ac{IDM} for reading-comprehension \ac{MCQ} is intrinsically hard. Difficulty depends on inferential demands across passage, question, and options jointly rather than on isolated lexical features, an empirical regularity that classical psychometric cognitive-component models of reading-comprehension items have characterised in detail \autocite{embretson1987component, gorin2006item, ozuru2008where}\rev{, and that computational text-analytic frameworks grounded in theories of comprehension operationalise through indices of referential and causal cohesion, syntactic simplicity, word concreteness and narrativity rather than through surface readability alone \autocite{graesser2011}}. Transformer fine-tuning without further modification thus leaves substantial room for improvement, unless one has access to large, high-quality datasets with diverse items (that are both content-diverse and combinatorially ``rich'', i.e., featuring multiple questions and options per passage), along with precise difficulty estimates \autocite{netik2024}. The recent shared task on automated prediction of item difficulty and item response time \autocite{yaneva2024} for medical \ac{MCQ} illustrates the regime: despite substantial model complexity across submissions, the best entries surpassed the so-called ``dummy'' regressor (a model that predicts the mean difficulty of the training set for every item) only marginally. The same small-sample pressure motivates recent work that integrates response-free predictions with response-based calibration as informative priors to reduce pilot-sample requirements \autocite{ulitzsch2025}; the response-free predictions themselves remain an essential input to that pipeline.

To the best of our knowledge, no published study has examined if and how different fine-tuning architectures can help in the \ac{IDM} task on reading-comprehension \ac{MCQ}s. Two questions remain open: (i) whether items are best encoded jointly, as a flat passage-question-options sequence, or component-wise; and (ii) whether the difficulty-regression target alone is enough, or whether adding an auxiliary task that more closely follows the process underlying test-taker behaviour (the mechanism that predominantly generates the difficulty) improves estimation, particularly in the small-sample regime where the labelled difficulty signal is sparse. \rev{Systematic manipulation of the training-set size has, to our knowledge, received little attention in the \ac{IDM} literature, and we are not aware of a comparable design on a dataset of this scale.}

To address these questions, we evaluate three transformer-based methods under a nested training-size design. \rev{The aim is not to maximise predictive accuracy, which would call for domain-specific pre-training and extensive hyper-parameter tuning, but to establish whether a fine-tuned transformer can predict difficulty from item wording without hand-engineered text features, and whether encoding the item's components separately or adding a second training objective improves it when training data are scarce.} A joint-encoding approach processes the concatenated item wording as a single input, collapsing the preprocessing and modelling stages of the feature-engineering tradition into one. A component-wise variant encodes the passage, question, and option set separately through a shared encoder before aggregation, making explicit at the architectural level the aforementioned functional decomposition of a \ac{MCQ}. A multi-task variant retains joint encoding and adds an auxiliary \ac{MCQA} objective sharing the encoder with the regression task; the auxiliary task forces option-level discrimination into the shared representation. Both extensions are motivated by the same concern: In the small-sample regime typical of applied psychometric work, transformer fine-tuning has limited information from which to recover task-relevant representations, and additional inductive biases (structural at the input level for component-wise encoding, supervisory at the loss level for multi-task learning) can act as forms of \textit{regularisation} that the regression objective on a scarce target alone cannot supply \autocite[or may not fully exploit,][]{geirhos2020}.

The remainder of the article proceeds as follows. Section~\ref{sec:methods} describes the data and pseudo-labelling procedure, defines the transformer-based approach with joint-encoding and the two encoding/training alternatives, and specifies the training and evaluation protocol. Section~\ref{sec:results} reports the empirical comparison across three training-set sizes. Section~\ref{sec:discussion} interprets the findings and discusses the limitations. Section~\ref{sec:conclusion} concludes the paper.

\section{Materials and methods}\label{sec:methods}

\subsection{Datasets}\label{sec:datasets}

\paragraph{Source corpus}
We work with the RACE++ collection \parencite[reading-comprehension multiple-choice items sourced from \rev{real} English-as-a-foreign-language exams in China;][]{liang2019}, balanced across the three source school types (middle, high, college) and across passages so that no school type or single passage is over-represented (full balancing recipe \rev{is disclosed} in supplementary materials), yielding a corpus of $29{,}618$ items. \rev{However, the Rasch difficulty estimate is available only for a subset of RACE++ which is published by \textcite{zou2022} as the ``PALRACE'' dataset.}

\rev{\paragraph{Pseudo-labelling procedure.}

\rev{To substantially expand the dataset for assessing the effect of dataset size and its potential interaction with different proposed methods, we needed to gather Rasch difficulty (i.e., the labels) for the rest of the items in the RACE++. Because collecting the human responses from which the difficulty is estimated, or obtaining expert ratings, is an immensely costly procedure, we resorted to so-called ``pseudo-labelling'' via a commercial \ac{LLM} and leveraged its few-shot learning capability \parencite[i.e., the ability of the model to learn from a handful of curated examples provided in the prompt]{brown2020}. Although this procedure is resource-intensive and depends on a closed-source model with publicly unknown architecture and parameter values (as opposed to the approaches provided in this article), we consider the method sufficiently performant and affordable for answering the paper's research questions. Specifically, we used Gemini 2.5 Flash via the official \ac{API} \parencite{comanici2025}, chosen on cost grounds. The purpose of the pseudo-labelling was to obtain a \textit{proxy} at least modestly correlated with the Rasch difficulty that we can use to demonstrate training dataset size sensitivity and to compare the proposed methods.
 
 The concrete procedure was as follows: the prompt for each target item (i.e. RACE++ item with unknown difficulty) draws one set of 20 PALRACE items with known Rasch difficulty, four from each quintile of the PALRACE difficulty distribution, so that the scale is represented evenly across its range. We pre-built 1{,}000 such sets and drew from them at random. At the end of the prompt, the target item wording is left without the difficulty label and the model is asked to fill the difficulty in. We used an \ac{API} feature called ``structured output'' that ensures the output is always numeric. Labels falling outside the interval $[-5, 5]$ were discarded as model errors, which removed approximately $80$ items; for reference, the PALRACE Rasch estimates range $[-4.4, 3.3]$.

 In pilot analyses preceding the full pseudo-labelling procedure, we experimented with several setups, every time using a held-out set of 50 PALRACE items. We varied the number of few-shot examples (number of items), their distribution (i.e. sampling based on quintiles vs. random sampling), or the presence of a system prompt (i.e. message consistently prepended to the varying prompt with PALRACE item wordings and target item wording). We also tried to keep the PALRACE ``few-shot'' sample set constant or to sample from a pool of pre-built sets. We evaluated the pseudo-labels' quality using a Spearman correlation coefficient against the true Rasch difficulty. The correlation ranged from low ($\rho = .21$) for prompts with a low number (i.e., five) of randomly (without stratification) sampled PALRACE items and with a detailed system prompt up to $\rho = .46$ for the procedure eventually used in this article. We found that the simple system prompt (\textit{``You are an expert in English comprehension test construction and psychometrics.''}) and random pre-built sets of 20 example items (each one with a headline, e.g., \textit{``EXAMPLE ITEM 15''} and \textit{``RASCH DIFFICULTY: +1.623''}) with the target item being marked as ``REAL ITEM'' worked the best. The configuration selected in these pilot analyses was then applied unchanged to the whole corpus: the same system prompt, the same output format, and the same example-sampling procedure were used for every item, with only the drawn set of 20 example items varying from item to item.}}

\rev{Applying the same procedure to the whole PALRACE dataset, each label formed without that item's own difficulty among its few-shot examples (i.e., in the leave-one-out manner), gives agreement of $\rho = .48$ (95\% CI $[.42, .53]$) with response-based Rasch difficulty across all 787 PALRACE items. This correlation bounds what any model trained on these pseudo-labels can reach. Because the training target from now onwards is the difficulty proxy and not the response-based estimate, a model that predicted the pseudo-labels perfectly would still correlate only about $.48$ with Rasch difficulty; we quantify how this ceiling propagates to the reported metrics in the Discussion.}

\paragraph{Train / validation / test partition.}
The pseudo-labelled corpus \rev{free of any PALRACE item} is partitioned at the \emph{passage} level into a training set $\mathcal{D}_{\text{train}}$ of $23{,}136$ items, a fixed validation set $\mathcal{D}_{\text{val}}$ of $n_{\text{val}} = 2{,}023$ items, and a held-out test set $\mathcal{D}_{\text{test}}$ of $4{,}459$ items; passage-level disjointness eliminates leakage through shared context. The partition is stratified on the cross of school type and the modal Rasch quintile of each passage's items.

\paragraph{Training sample sizes.}
From the training set $\mathcal{D}_{\text{train}}$ we draw nested sub-samples that vary training sample size by Monte Carlo subsampling. We consider three sizes, $n \approx 800$, $n \approx 4{,}300$, and $n \approx 23{,}000$, spaced roughly logarithmically; the largest equals $\mathcal{D}_{\text{train}}$ exactly. For each size $n$ and each of ten sample seeds $s \in \{42, \ldots, 51\}$ we draw a sub-sample $\mathcal{T}_{n,s} \subset \mathcal{D}_{\text{train}}$ as a passage-level prefix of a per-seed shuffle, stratified on the same factors as the corpus partition. The same shuffle is reused across the three sizes, so that for each fixed seed, the sub-samples are nested,
\begin{equation}
  \label{eq:nested}
  \mathcal{T}_{n,s} \,\subset\, \mathcal{T}_{n',s} \quad \text{for } n < n',
\end{equation}
while across-seed draws are independent. $\mathcal{D}_{\text{val}}$ and $\mathcal{D}_{\text{test}}$ are fixed across all cells, isolating the effect of training-set size from sampling noise.

\subsection{General modelling framework}\label{sec:framework}

The encoder-only family of transformers \citep[\ac{BERT} and its successors;][]{devlin2019} consists of stacks of self-attention layers \citep{vaswani2017} pre-trained on a general-purpose corpus by \ac{MLM} \autocite[for the linguistic regularities encoded by such pre-training, see][]{jawahar2019}. \rev{In every self-attention layer, the representation of each token is recomputed as a learned weighted combination of the representations of all tokens in the input, so that each part of the wording is encoded in the context of the entire item rather than in isolation.} The \emph{encoder-only} qualifier distinguishes this family from the autoregressive \emph{decoder-only} models used for free-form text generation (such as the \ac{LLM} employed in Section~\ref{sec:datasets} for pseudo-labelling): both rely on the same self-attention mechanism but differ in pre-training objective and use case, with encoder-only models optimised to output representations suitable for downstream prediction tasks rather than to produce text. For fine-tuning to a particular task, the \ac{MLM} portion of the pre-trained model can be removed to form a ``body'' and any small task-specific ``head'' appended. In our case, the head is denoted $g_\psi$ and maps a pooled representation of the text input to the item difficulty; the methods differ in the input given to $g_\psi$ and in its form (Sections~\ref{sec:baseline}--\ref{sec:mtl}), and in whether an auxiliary head and auxiliary objective are present alongside the regression (Section~\ref{sec:mtl}).

The general setup is as follows: a pre-trained encoder-only transformer, denoted $\operatorname{Enc}_\phi$, maps an input token sequence $\mathbf{x}_i$ (not necessarily equal in length to the number of words, see below) for input $i$ to its last-layer hidden states $\mathbf{h}_i$,
\begin{equation*}
  \mathbf{h}_i \,=\, \operatorname{Enc}_\phi(\mathbf{x}_i) \,\in\, \mathbb{R}^{L \times d},
\end{equation*}
where $L$ is the input length, $d$ is the encoder hidden size (\rev{i.e., the number of features the model uses to represent each token;} 768 for the \texttt{ModernBERT-base} model that we use here; \citealp{warner2024}), and $\phi$ collects the encoder's trainable parameters (149 million for \texttt{ModernBERT-base}). The tokeniser (i.e., the algorithm that splits the input into sub-word tokens) prepends a special \texttt{[CLS]} token introduced in \ac{BERT} \parencite{devlin2019} as the first token of every input, which is pre-trained to summarise the sequence via a pooled representation; we write $\mathbf{h}_{[\text{CLS}]} \in \mathbb{R}^d$ for its last-layer hidden state and use it as the pooled sequence representation that the head $g_\psi$ receives,
\begin{equation*}
  \mathbf{r}_i \,=\, (\mathbf{h}_i)_{[\text{CLS}]} \,\in\, \mathbb{R}^d.
\end{equation*}

The encoder parameters~$\phi$ and the head parameters~$\psi$ are trained \textit{jointly} by gradient descent using the AdamW optimiser \citep{loshchilov2019decoupled}, as implemented in the training loop of the \texttt{transformers} Python framework \parencite{wolf2020transformers}, by minimising a task-specific loss~$\mathcal{L}$ defined separately for each of the three methods (Sections~\ref{sec:baseline}--\ref{sec:mtl}). Optimisation hyper-parameters are held fixed across all methods and training sizes; we defer their values to  Section~\ref{sec:training} so that all training-time choices are gathered in one place.

Throughout this paper, $b_i$ denotes the difficulty of item~$i$ in the Rasch model \citep{rasch1960}. In our case, $b_i$ is an \ac{LLM} pseudo-label based on difficulty estimates from human response patterns to RACE++ items included in the PALRACE dataset (see Section~\ref{sec:datasets} for further details). The model's text-based prediction is denoted $\hat b_i$.

\subsection{Joint-encoding}\label{sec:baseline}

The joint-encoding approach follows the general transfer-learning approach and fine-tuning procedure established in the \ac{NLP} field by \citet{howard2018finetuning}, and serves as the reference point against which the two methodological extensions are compared, along with the ``dummy'' regressor baseline. The complete item wording is (i) concatenated and (ii) tokenised into a single input sequence $\mathbf{x}_i$ consisting of the passage, question, and all options of item~$i$, with the special \texttt{[SEP]} tokens introduced in \ac{BERT} \parencite{devlin2019} marking the boundaries between individual components and the \texttt{[CLS]} token prepended at the beginning of the sequence. The encoder produces the pooled representation $\mathbf{r}_i = (\operatorname{Enc}_\phi(\mathbf{x}_i))_{[\text{CLS}]}$ defined in Section~\ref{sec:framework}, and a linear layer (the head equivalent to multiple linear regression) maps it directly to the predicted difficulty,
\begin{equation}
  \label{eq:baseline}
  \hat{b}_i \,=\, g_\psi(\mathbf{r}_i) \,=\, \mathbf{w}^\top \mathbf{r}_i + c,
\end{equation}
where $\mathbf{w} \in \mathbb{R}^d$ is the head's weight vector and $c \in \mathbb{R}$ its scalar bias, both learned from data alongside the encoder parameters~$\phi$ (which are not initialised \textit{de novo} but retained from the intensive pre-training). In this form, the pooled encoder output $\mathbf{r}_i$ can be understood as a learned feature vector of the item and $\mathbf{w}$ as the corresponding feature weights, both optimised from data rather than specified by the researcher. Equation~\eqref{eq:baseline} has the same linear form as the \ac{LLTM} referenced in the Introduction, but with one key difference -- the feature vector $\mathbf{r}_i$ is not fixed; the encoder parameters~$\phi$ are optimised jointly with~$\mathbf{w}$, so that both the features and their weights are learned from data.

  The joint-encoding model is trained by minimising the \ac{MSE} between the predicted and target difficulties,
  \begin{equation}
    \label{eq:loss-reg}
    \mathcal{L}_{\text{reg}}
    \,=\, \frac{1}{N}\sum_{i=1}^{N}
    \bigl(\hat{b}_i - b_i\bigr)^2.
  \end{equation}
  The same \ac{MSE} loss is used by the component-wise variant in Section~\ref{sec:encoding}; the multi-task variant in Section~\ref{sec:mtl} retains it and adds a second loss term for the auxiliary task.

The following Sections~\ref{sec:encoding} and~\ref{sec:mtl} describe two extensions of this approach: component-wise encoding, which modifies the input to $\operatorname{Enc}_\phi$ so that item components are processed separately before aggregation, and \ac{MTL}, which adds an auxiliary objective alongside the \ac{MSE} for the regression task.

\begin{figure}[t]
  \centering
  \begin{subfigure}[b]{0.135\textwidth}
    \centering
    \resizebox{\textwidth}{!}{\begin{tikzpicture}[
          node distance=3mm and 5mm,
          block/.style = {draw, rounded corners=2pt, minimum width=22mm, minimum height=6mm, align=center, font=\footnotesize, inner sep=3pt},
          enc/.style   = {draw, fill=gray!12, rounded corners=2pt, minimum width=22mm, minimum height=8mm, align=center, font=\footnotesize, inner sep=3pt},
          inp/.style   = {draw, dashed, minimum width=22mm, minimum height=7mm, align=center, font=\scriptsize, inner sep=2pt},
          outl/.style  = {font=\footnotesize\itshape},
          arr/.style   = {-{Latex[length=1.6mm]}, semithick},
        ]
        \node[inp] (a-in) {passage \textnormal{[SEP]}\\ question \textnormal{[SEP]}\\ options};
        \node[enc,   above=of a-in]    (a-enc)  {$\operatorname{Enc}_\phi$};
        \node[block, above=of a-enc]   (a-cls)  {$\mathbf{r}_i$};
        \node[block, above=of a-cls]   (a-head) {regression $g_\psi$\\[-1pt]{\scriptsize\itshape linear}};
        \node[outl,  above=of a-head]  (a-out)  {$\hat b_i$};
        \draw[arr] (a-in)  -- (a-enc);    \draw[arr] (a-enc)  -- (a-cls);
        \draw[arr] (a-cls) -- (a-head);   \draw[arr] (a-head) -- (a-out);
    \end{tikzpicture}}
    \caption{Joint encoding}
    \label{fig:methods-a}
  \end{subfigure}\hfill
  \begin{subfigure}[b]{0.42\textwidth}
    \centering
    \resizebox{\textwidth}{!}{\begin{tikzpicture}[
          node distance=3mm and 5mm,
          block/.style = {draw, rounded corners=2pt, minimum width=22mm, minimum height=6mm, align=center, font=\footnotesize, inner sep=3pt},
          enc/.style   = {draw, fill=gray!12, rounded corners=2pt, minimum width=22mm, minimum height=8mm, align=center, font=\footnotesize, inner sep=3pt},
          inp/.style   = {draw, dashed, minimum width=22mm, minimum height=7mm, align=center, font=\scriptsize, inner sep=2pt},
          outl/.style  = {font=\footnotesize\itshape},
          arr/.style   = {-{Latex[length=1.6mm]}, semithick},
        ]
        \node[inp] (b-in2) {question};
        \node[inp, left=1.5mm  of b-in2] (b-in1) {passage};
        \node[inp, right=1.5mm of b-in2] (b-in3) {options};
        \node[enc,   above=5mm of b-in2, minimum width=68mm] (b-enc) {shared $\operatorname{Enc}_\phi$};
        \node[block, anchor=south, minimum width=18mm] (b-cls1) at ($(b-enc.north -| b-in1) + (0,5mm)$) {$\mathbf{r}_i^{(1)}$};
        \node[block, anchor=south, minimum width=18mm] (b-cls2) at ($(b-enc.north -| b-in2) + (0,5mm)$) {$\mathbf{r}_i^{(2)}$};
        \node[block, anchor=south, minimum width=18mm] (b-cls3) at ($(b-enc.north -| b-in3) + (0,5mm)$) {$\mathbf{r}_i^{(3)}$};
        \node[draw, ellipse, align=center, font=\footnotesize, inner xsep=5pt, inner ysep=1pt, above=5mm of b-cls2] (b-cat) {concat};
        \node[block, above=4mm of b-cat] (b-head) {regression $g_\psi$\\[-1pt]{\scriptsize\itshape two-layer, GELU}};
        \node[outl,  above=of b-head] (b-out)  {$\hat b_i$};
        \draw[arr] (b-in1.north) -- (b-in1.north |- b-enc.south);
        \draw[arr] (b-in2.north) -- (b-in2.north |- b-enc.south);
        \draw[arr] (b-in3.north) -- (b-in3.north |- b-enc.south);
        \draw[arr] (b-cls1.south |- b-enc.north) -- (b-cls1.south);
        \draw[arr] (b-cls2.south |- b-enc.north) -- (b-cls2.south);
        \draw[arr] (b-cls3.south |- b-enc.north) -- (b-cls3.south);
        \draw[arr] (b-cls1.north) -- (b-cat);
        \draw[arr] (b-cls2.north) -- (b-cat);
        \draw[arr] (b-cls3.north) -- (b-cat);
        \draw[arr] (b-cat) -- (b-head);
        \draw[arr] (b-head) -- (b-out);
    \end{tikzpicture}}
    \caption{Component-wise encoding}
    \label{fig:methods-b}
  \end{subfigure}\hfill
   \begin{subfigure}[b]{0.36\textwidth}
      \centering
      \resizebox{\textwidth}{!}{\begin{tikzpicture}[
            node distance=3mm and 5mm,
            block/.style = {draw, rounded corners=2pt, minimum width=22mm, minimum height=6mm, align=center, font=\footnotesize, inner sep=3pt},
            enc/.style   = {draw, fill=gray!12, rounded corners=2pt, minimum width=22mm, minimum height=8mm, align=center, font=\footnotesize, inner sep=3pt},
            inp/.style   = {draw, dashed, minimum width=22mm, minimum height=7mm, align=center, font=\scriptsize, inner sep=2pt},
            outl/.style  = {font=\footnotesize\itshape},
            arr/.style   = {-{Latex[length=1.6mm]}, semithick},
          ]
          \node[inp, minimum width=22mm, align=center] (c-in-r) {passage \textnormal{[SEP]}\\ question \textnormal{[SEP]}\\ options};
          \node[inp, right=4mm of c-in-r, minimum width=26mm, align=center] (c-in-m) {$\bigl[\text{passage}\,;\,\text{question}\,;\,\text{option-}m\bigr]$};
          \node[enc, minimum width=52mm, anchor=south, align=center] (c-enc) at ($(c-in-r.north)!0.5!(c-in-m.north) + (0,5mm)$) {shared
          $\operatorname{Enc}_\phi$\\[-1pt]{\scriptsize\itshape +\,$\mathbf{z}_t$ at \texttt{[CLS]}}};
          \node[block, anchor=south, minimum width=22mm] (c-cls-r) at ($(c-enc.north -| c-in-r) + (0,5mm)$) {$\mathbf{r}_i$};
          \node[block, anchor=south, minimum width=26mm] (c-cls-m) at ($(c-enc.north -| c-in-m) + (0,5mm)$) {$\mathbf{r}_i^{(\text{opt}_m)}$};
          \node[block, minimum width=22mm, above=4mm of c-cls-r] (c-headA) {reg.\ $g_\psi$};
          \node[block, minimum width=26mm, above=4mm of c-cls-m] (c-headB) {clf. with $f_\xi$};
          \node[outl, above=of c-headA] (c-pred-b) {$\hat b_i$};
          \node[outl, above=of c-headB] (c-pred-m) {$p_{i,m}$};
          \draw[arr] (c-in-r.north) -- (c-in-r.north |- c-enc.south);
          \draw[arr] (c-in-m.north) -- (c-in-m.north |- c-enc.south);
          \draw[arr] (c-cls-r.south |- c-enc.north) -- (c-cls-r.south);
          \draw[arr] (c-cls-m.south |- c-enc.north) -- (c-cls-m.south);
          \draw[arr] (c-cls-r.north) -- (c-headA.south);
          \draw[arr] (c-cls-m.north) -- (c-headB.south);
          \draw[arr] (c-headA.north) -- (c-pred-b.south);
          \draw[arr] (c-headB.north) -- (c-pred-m.south);
          \node[font=\scriptsize\itshape, align=center, anchor=south] (c-altlbl) at
          ($(c-pred-b.north)!0.5!(c-pred-m.north) + (0,6mm)$) {tasks alternated per batch\\(random scheduler)};
          \draw[densely dashed, thick]
            (c-pred-b.north) -- (c-pred-b.north |- c-altlbl.south)
            -- (c-pred-m.north |- c-altlbl.south) -- (c-pred-m.north);
      \end{tikzpicture}}
      \caption{Multi-task learning}
      \label{fig:methods-c}
    \end{subfigure}
  \caption{Three response-free item-difficulty models compared in this work. (a)~The \emph{joint-encoding} approach encodes the concatenated wording and predicts
    $\hat b_i$ from the \texttt{[CLS]} token hidden state. (b)~The \emph{component-wise} variant encodes passage, question, and options separately through the
    shared encoder $\operatorname{Enc}_\phi$ and concatenates the three \texttt{[CLS]} vectors before regression. (c)~The \emph{multi-task} model retains joint
    encoding and adds a \ac{MCQA} classification branch that shares the encoder with the regression branch; the regression head additionally drives the
    prediction $\hat b_i$. A learnable task-conditioning vector $\mathbf{z}_t$ is added at the \texttt{[CLS]} position of the embedding-layer output (see
  Figure~\ref{fig:taskcls}). The two branches alternate per batch under a random task scheduler; at each step, only the branch chosen for that batch is active, so the regression branch produces $\hat b_i$ on
  regression steps and the MCQA branch predicts the probability that option $m$ for item $i$ is correct: $p_{i,m} = \exp(z_{i,m})/ \sum_{k=1}^{M} \exp(z_{i,k})$ on MCQA steps.}
  \label{fig:methods}
\end{figure}

\subsection{Component-wise encoding}
\label{sec:encoding}

An \ac{MCQ} has a natural functional decomposition into passage, question, and option set \autocite{embretson1987component}, and a principled way to make this structure explicit at the architectural level is to encode each component separately through a shared encoder and aggregate the resulting representations before regression. In the joint-encoding approach, standard self-attention treats all input tokens symmetrically, so no architectural signal distinguishes which tokens belong to which functional component, and the model must learn this segmentation implicitly from formatting cues such as option markers \citep[for how self-attention heads acquire syntactic and structural regularities during pre-training, see][]{rogers2020primer}.

A second motivation is that the difficulty-relevant properties of distractors are fundamentally relational: a distractor is attractive not because of its surface form in isolation, but because of its semantic proximity to the correct answer or its plausibility given the question. A single-vector pooling of a flat input asks the encoder to preserve such relational features alongside all other item characteristics through one bottleneck, a demand that grows with item length. By encoding components separately and combining them later, the component-wise variant retains per-component representations and gives the regression head access to a richer, more structured input.

Instead of processing the entire item as a flat sequence, the component-wise variant encodes each functional component of the item separately. Let $\mathbf{x}_i^{(j)}$ denote the token sequence of the $j$-th component of item~$i$, with $j = 1, \ldots, C$ indexing the passage, the question, and the options. We adopt $C = 3$ and join the four answer options into a single labelled string for the third component (with \texttt{A)}, \texttt{B)}, \texttt{C)}, and \texttt{D)} prepended to each option -- as in the case of the joint-encoding approach above).

Each component is passed through the shared encoder, yielding last-layer hidden states $\mathbf{h}_i^{(j)} = \operatorname{Enc}_\phi(\mathbf{x}_i^{(j)})$, and is represented by its \texttt{[CLS]} hidden state, in direct analogy with the joint-encoding construction of $\mathbf{r}_i$:
\begin{equation}
\label{eq:component}
\mathbf{r}_i^{(j)} \,=\, \bigl(\mathbf{h}_i^{(j)}\bigr)_{[\text{CLS}]}
\,\in\, \mathbb{R}^d, \qquad j = 1, \ldots, C.
\end{equation}
The encoder parameters~$\phi$ are shared across components: the same weights process the passage, the question, and the options. The three component representations are concatenated along the feature axis \rev{into a single vector $\mathbf{z}_i$:} 
\rev{
$$\mathbf{z}_i \,=\, \bigl[\mathbf{r}_i^{(1)};\; \ldots\;;\; \mathbf{r}_i^{(C)}\bigr],$$
}
\rev{where $[\,\cdot\,;\,\cdot\,]$ denotes concatenation. The vector $\mathbf{z}_i$ is then passed to the head $g_\psi$ to arrive at the predicted difficulty:}
\begin{equation}
\label{eq:structured}
\hat{b}_i \,=\, g_\psi\!(\mathbf{z}_i), \qquad g_\psi \colon \mathbb{R}^{C \cdot d} \to \mathbb{R}.
\end{equation}
\rev{The form of $g_\psi$ is more expressive here than in the case of the joint-encoding approach. Instead of a simple linear layer, we introduce a small two-layer network with an activation function in between, so the vectors representing distinct wording components can interact in a non-linear fashion:
\begin{equation}
\label{eq:structured-head}
g_\psi(\mathbf{z}_i) \,=\, \mathbf{w}^\top \sigma(\mathbf{W}\mathbf{z}_i + \mathbf{b}) + c,
\end{equation}
where $\mathbf{W} \in \mathbb{R}^{d \times C \cdot d}$ and $\mathbf{b} \in \mathbb{R}^{d}$ are the weight matrix and bias of the aggregation layer which projects the original concatenated vector of length $C \cdot d$ to the hidden size $d$, $\sigma$ is the \ac{GELU} activation function, and $\mathbf{w} \in \mathbb{R}^{d}$, $c \in \mathbb{R}$ form the final linear layer leading to the difficulty scalar.}
The training objective is unchanged from the joint-encoding version, $\mathcal{L}_{\text{reg}}$ in Equation~\eqref{eq:loss-reg}.

\subsection{Multi-task learning}
\label{sec:mtl}

The \ac{MTL} extension addresses a complementary gap. Even with component-wise encoding, the model is trained solely to minimise prediction error on a scalar difficulty target, which provides no signal about \emph{why} an item is easy or hard. Adding an auxiliary question-answering objective forces the encoder to develop representations that enable option-level discrimination, that is, representations from which the correct answer can be distinguished from distractors. To form this representation, the model must ``attend'' to the passage, the question and each option in the way required to choose the key. In the terminology of \citet{ruder2017overview} (building on the original formulation of \citealp{caruana1997}), the auxiliary task provides an \textit{inductive bias} that causes the model to prefer representations useful for more than one objective, acting as a form of \textit{regularisation} on the \textit{shared encoder} that is most consequential when the primary regression signal is sparse or offers surface-level shortcuts (such as attending only to the passage and ignoring the rest of the wording), that is, in the small-sample regime. Here, the bias is also substantively and psychologically motivated: to solve the item, a test-taker must evaluate each option against the question and passage, and the \ac{MTL} objective encourages the model to perform an analogous computation. In this sense, the auxiliary task is not only a regulariser: it also aligns the learned features with the \textit{process that generates difficulty} in the first place. \rev{Answering ability is not itself the difficulty estimate: the prediction comes directly from the regression head $g_\psi$ trained on the difficulty target, not from the model's success or confidence in answering. However, the auxiliary gradient does reach the shared encoder, and shaping that representation is its intended role.}

For the \ac{MCQA} task, a custom input stemming from the same item is presented $M$ times, once per answer option $m$ (for $m = 1, \ldots, M$; with $M = 4$ in our case). For each option, the shared context $[\text{[CLS]}; \text{passage}; \text{question}; \text{[SEP]}]$ is concatenated with $\text{option}_m$ to form the input token sequence $\mathbf{x}_i^{(\text{opt}_m)}$, and its pooled representation $\mathbf{r}_i^{(\text{opt}_m)} = \bigl(\operatorname{Enc}_\phi(\mathbf{x}_i^{(\text{opt}_m)})\bigr)_{[\text{CLS}]}$ is the \texttt{[CLS]} hidden state of the encoder's output for that input. Each pooled representation is passed through a classification head $f_\xi$ that produces a scalar ``logit''\footnote{\rev{``Logit'' is used here in the machine-learning sense: the unnormalised score assigned to a response option before the softmax converts it into a probability. It is not the logit unit of the latent scale in \ac{IRT}.}}.
The function $f_\xi$ has the same form as $g_\psi$ of Equation~\eqref{eq:baseline}, with its own parameters collected under $\xi = \{\mathbf{u} \in \mathbb{R}^d,\, e \in \mathbb{R}\}$, shared across all $M$ options (input sequences):
\begin{equation}
\label{eq:mcqa-logit}
z_{i,m} \,=\, f_\xi\!\bigl(\mathbf{r}_i^{(\text{opt}_m)}\bigr) \,=\, \mathbf{u}^\top \mathbf{r}_i^{(\text{opt}_m)} + e
\qquad \text{for } m = 1, \ldots, M\,.
\end{equation}
What distinguishes the two heads is the task they serve and the loss they are paired with: $g_\psi$ outputs a scalar prediction trained under the \ac{MSE} loss (Equation~\ref{eq:loss-reg}); $f_\xi$ outputs per-option ``logits''~$z_{i,m}$ and its parameters~$\xi$ are trained using the cross-entropy loss defined below (Equation~\ref{eq:mcqa-loss}). Keeping $f_\xi$ as a single linear layer devotes the discriminative capacity required to distinguish correct from incorrect options to the encoder representations themselves, which is the reason we employ this auxiliary task in the first place -- to constrain the representations for the main \ac{IDM} goal.

The \ac{MCQA} loss is calculated using standard cross-entropy, where the model's predicted raw logits over the $M$ options are passed through a softmax function (in machine-learning terminology, equivalent to multinomial logistic regression), and the negative log-likelihood is computed only with respect to the integer index $m_{i}^{*}$ of the correct option:
\begin{equation}
\label{eq:mcqa-loss}
\mathcal{L}_{\text{MCQA}}
= -\frac{1}{N}\sum_{i=1}^{N}
\log \frac{\exp(z_{i,m^*_i})}
{\sum_{m=1}^{M} \exp(z_{i,m})}\,.
\end{equation}

Training \textit{alternates} between the two tasks: the data loader interleaves equal numbers of batches per task, so that over a full training epoch each task contributes the same number of optimiser steps; the shared encoder parameters~$\phi$ are updated at every step. The regression head parameters~$\psi$ receive ``gradients'' only from $\mathcal{L}_{\text{reg}}$, and the \ac{MCQA} head parameters~$\xi$ only from $\mathcal{L}_{\text{MCQA}}$, while the shared encoder~$\phi$ is updated by whichever task is active in a given step. We do not form a joint loss per batch.

A practical concern in \ac{MTL} is so-called \emph{negative transfer} \autocite{wang2019negativetransfer}, where the gradients of one task interfere destructively with those of the other and degrade the shared representation rather than enrich it. Two design choices in our setup seek to mitigate this concern. First, we weight the \ac{MCQA} loss to keep it from dominating the regression loss on the shared encoder: the regression \ac{MSE} and the cross-entropy may not be balanced in terms of scale, and a plain sum can let the auxiliary task dictate the updates. We multiply the \ac{MCQA} loss by a constant $\lambda = 0.05$ \autocite{gong2019mtlweights}, selected by grid search over $\{0.005, 0.05, 0.5, 1\}$ on a separate \rev{$n = 799$} sample from the training set that is not used in any sub-sample in this study \rev{(validation RMSE was 1.119 at $\lambda = 0.005$, 1.106 at $\lambda = 0.05$, 1.121 at $\lambda = 0.5$, and 1.130 at $\lambda = 1$)}.
Because our implementation employs a stochastic task scheduler rather than computing a joint loss per batch, the gradient $\mathbf{g}_s$ backpropagated to the shared encoder at step $s$ is either $\nabla\mathcal{L}_{\text{reg}}$ or $\lambda\nabla\mathcal{L}_{\text{MCQA}}$. As tasks are sampled with equal probability, the expected gradient update is:

$$ \mathbb{E}[\mathbf{g}_s] = \frac{1}{2}\bigl(\nabla\mathcal{L}_{\text{reg}} + \lambda\nabla\mathcal{L}_{\text{MCQA}}\bigr) $$

Second, to give the shared encoder an explicit signal about which task is being processed at each step, we introduce a learnable task-conditioning matrix $\mathbf{Z} \in \mathbb{R}^{T \times d}$, following the additive input-level conditioning scheme used by \citet{sileo2023tasksource}, in this case only with $T = 2$ rows (one per task). The encoder hidden size is $d$. During the forward pass for the active task $t \in \{\text{regression}, \text{MCQA}\}$, the row $\mathbf{z}_t \in \mathbb{R}^d$ is added to the embedding-layer output at the \texttt{[CLS]} position before the first encoder layer reads the representation,
\begin{equation}
\label{eq:taskcls}
\tilde{\mathbf{e}}_{i,\,[\text{CLS}]}
\,=\, \mathbf{e}_{i,\,[\text{CLS}]} + \mathbf{z}_t,
\end{equation}
where $\mathbf{e}_{i,\,[\text{CLS}]}$ is the pre-conditioning embedding of the \texttt{[CLS]} token. $\mathbf{Z}$ is initialised to zero, so training begins without any conditioning; the rows acquire non-zero values only to the extent that the shared encoder benefits from a task-specific adjustment. Figure~\ref{fig:taskcls} illustrates the placement of $\mathbf{z}_t$.

\begin{figure}[t]
\centering
\begin{tikzpicture}[
node distance=6mm,
stage/.style    = {draw, rounded corners=2pt, minimum width=30mm, minimum height=8mm, align=center, font=\footnotesize},
inj/.style      = {draw, fill=gray!12, rounded corners=2pt, minimum width=26mm, minimum height=9mm, align=center, font=\footnotesize},
extern/.style   = {draw, dashed, rounded corners=2pt, minimum width=22mm, minimum height=8mm, align=center, font=\footnotesize},
arr/.style      = {-{Latex[length=1.6mm]}, semithick},
]
\node[stage] (emb) {embedding layer};
\node[inj,   right=of emb] (add) {$+\,\mathbf{z}_t$\\ at \texttt{[CLS]}};
\node[stage, right=of add] (enc) {1\textsuperscript{st} encoder  layer};
\draw[arr] (emb) -- (add);
\draw[arr] (add) -- (enc);

\node[extern, above=10mm of add] (Z) {$\mathbf{Z} \in \mathbb{R}^{T\times d}$};
\draw[arr, dashed] (Z) -- node[right, font=\scriptsize] {select row $t$} (add);

\end{tikzpicture}
\caption{Task-conditioning vector $\mathbf{z}_t$ in the multi-task model. The matrix $\mathbf{Z} \in \mathbb{R}^{T\times d}$ stores one learnable vector per task ($T = 2$). During a forward pass, the active task selects row $\mathbf{z}_t$, which is added only at the \texttt{[CLS]} position of the embedding-layer output before the first encoder layer. Both rows of $\mathbf{Z}$ are initialised to zero and trained jointly with the rest of the model.}
\label{fig:taskcls}
\end{figure}

\subsection{Training procedure and evaluation}
\label{sec:training}

A single fine-tuning regime is applied to every (training sample size $\times$ method $\times$ sub-sample) cell of the experimental design, so that within-size differences between methods reflect the architectural choice rather than hyper-parameter selection (although we recognise that a specific hyper-parameter setup may further improve the performance of each method). We optimise with AdamW \autocite{loshchilov2019decoupled} under a fixed schedule and mini-batch configuration (exact values in supplementary materials), stop training early based on the validation \ac{RMSE}, and restore the best-on-validation checkpoint for the final test evaluation. The held-out test split $\mathcal{D}_{\text{test}}$ is consulted only once per cell, after training has finished. Final metrics are \ac{RMSE} (primary), Spearman $\rho$, and $R^2$; at $n \approx 800$ and $n \approx 4{,}300$ we report mean and standard deviation across the ten sample seeds (Section~\ref{sec:datasets}) to expose the dependence of each metric on the training sub-sample drawn, and a single value at $n \approx 23{,}000$ (as we cannot subsample the whole training set in any way). A \emph{constant} ``dummy'' regressor that returns the training-set mean of $b_i$ for every test item is evaluated at every cell as a model-free reference, following the convention of \citet{yaneva2024}. \rev{The source corpora are publicly available from their original publications (Section~\ref{sec:datasets}), and the detailed dataset preparation description, hyper-parameters, and other technical specifics are provided in the supplementary materials deposited on the Open Science Framework (OSF).}

\section{Results}\label{sec:results}

We present the results in two tables and three figures. Table~\ref{tab:main-results} reports absolute test-set performance per training-set size and method. Table~\ref{tab:wilcox} reports paired two-sided Wilcoxon signed-rank tests of within-subsample differences against joint-encoding, across the twelve cells defined by the two comparator methods, two training sizes, and three metrics.

Three results stand out. First, at $n \approx 800$ \ac{MTL} improves on joint-encoding across all three metrics, with 95\%~confidence intervals on the Hodges--Lehmann median shift excluding zero and 9 of 10 sub-samples favouring \ac{MTL} for each metric. Second, the improvement is no longer detectable at $n \approx 4{,}300$ (all CIs straddle zero, 5--6 of 10 sub-samples favour \ac{MTL}), consistent with the design hypothesis that the auxiliary objective regularises most when the labelled difficulty signal is sparse. Third, the component-wise variant is statistically indistinguishable from joint-encoding on \ac{RMSE} and $R^2$ at both training sizes (all CIs include zero), and improves the rank ordering of items at $n \approx 800$ (Spearman $\rho$: $\hat{\Delta}_{\text{HL}} = 0.016$, 95\%~CI $[0.005, 0.025]$, 9 of 10 sub-samples favourable), an advantage that, like the \ac{MTL} one, is no longer detectable at $n \approx 4{,}300$.

\begin{table}[h!]
\centering
\caption{Test-set performance per training-set size and method}
\label{tab:main-results}
\vspace{4pt}
\renewcommand{\arraystretch}{1.15}
\small
\sbox0{\begin{tabular}{r l c c c}
\toprule
$n$ & Method & RMSE & $R^2$ & Spearman $\rho$ \\
\midrule
$\approx 800$     & Dummy reg.     & $1.299\;(0.000)$ & $.000\;(0.000)$ & n/a \\
                  & Joint-encoding & $1.101\;(0.013)$ & $.281\;(0.017)$ & $.535\;(0.015)$ \\
                  & Component-wise & $1.099\;(0.012)$ & $.284\;(0.016)$ & $.550\;(0.005)$ \\
                  & \acs{MTL}      & $\mathbf{1.088}\;(0.009)$ & $\mathbf{.299}\;(0.011)$ & $\mathbf{.551}\;(0.011)$ \\
\midrule
$\approx 4{,}300$ & Dummy reg.     & $1.299\;(0.000)$ & $.000\;(0.000)$ & n/a \\
                  & Joint-encoding & $1.053\;(0.007)$ & $.342\;(0.009)$ & $.598\;(0.008)$ \\
                  & Component-wise & $1.053\;(0.010)$ & $.343\;(0.013)$ & $.593\;(0.011)$ \\
                  & \acs{MTL}      & $\mathbf{1.049}\;(0.010)$ & $\mathbf{.348}\;(0.012)$ & $\mathbf{.600}\;(0.009)$ \\
\midrule
$\approx 23{,}000$ & Dummy reg.     & $1.299$ & $.000$ & n/a \\
                   & Joint-encoding & $1.010$ & $.396$ & $.635$ \\
                   & Component-wise & $1.023$ & $.380$ & $.623$ \\
                   & \acs{MTL}      & $\mathbf{1.003}$ & $\mathbf{.404}$ & $\mathbf{.645}$ \\
\bottomrule
\end{tabular}}\begin{minipage}{\wd0}
\usebox0\par
\vspace{4pt}
\footnotesize \textit{Note.} Values are means across ten random sample seeds at $n \approx 800$ and $n \approx 4{,}300$, with standard deviations in parentheses; a single run on the full training set at $n \approx 23{,}000$. The dummy reg. baseline (Section~\ref{sec:training}) returns the training-set mean of $b_i$ for every test item; its Spearman $\rho$ is undefined because its predictions have zero variance. The best entry per metric within each training-set size is set in bold.
\end{minipage}
\end{table}

\begin{table}[h!]
  \centering
  \caption{Paired Wilcoxon signed-rank tests of within-sub-sample differences against joint-encoding}
  \label{tab:wilcox}
  \vspace{4pt}
  \renewcommand{\arraystretch}{1.15}
  \small
  \sbox0{\begin{tabular}{l r l c c c c c}
  \toprule
  Method & $n$ & Metric & $\hat{\Delta}_{\text{HL}}$ & 95\% CI & $V$ & $p$ & favourable \\
  \midrule
  Component-wise & $\approx 800$    & \ac{RMSE}      & $-0.002$ & $[-0.013, 0.006]$ & $23$ & $.695$ & $5/10$ \\
                 &                  & $R^2$          & $0.003$ & $[-0.008, 0.017]$ & $32$ & $.695$ & $5/10$ \\
                 &                  & Spearman $\rho$ & $0.016$ & $[0.005, 0.025]$ & $50$ & $\mathbf{.020}$ & $9/10$ \\
  \addlinespace
  & $\approx 4{,}300$ & \ac{RMSE}      & $-0.002$ & $[-0.011, 0.011]$ & $23$ & $.695$ & $6/10$ \\
                 &                   & $R^2$          & $0.003$ & $[-0.013, 0.014]$ & $32$ & $.695$ & $6/10$ \\
                 &                   & Spearman $\rho$ & $-0.005$ & $[-0.016, 0.005]$ & $16$ & $.275$ & $3/10$ \\
  \midrule
  \acs{MTL} & $\approx 800$    & \ac{RMSE}      & $-0.013$ & $[-0.021, -0.006]$ & $1$  & $\mathbf{.004}$ & $9/10$ \\
            &                  & $R^2$          & $0.017$ & $[0.008, 0.027]$ & $54$ & $\mathbf{.004}$ & $9/10$ \\
            &                  & Spearman $\rho$ & $0.016$ & $[0.004, 0.023]$ & $53$ & $\mathbf{.006}$ & $9/10$ \\
  \addlinespace
   & $\approx 4{,}300$ & \ac{RMSE}      & $-0.005$ & $[-0.013, 0.004]$ & $18$ & $.375$ & $5/10$ \\
            &                   & $R^2$          & $0.006$ & $[-0.005, 0.016]$ & $37$ & $.375$ & $5/10$ \\
            &                   & Spearman $\rho$ & $0.003$ & $[-0.007, 0.011]$ & $35$ & $.492$ & $6/10$ \\
  \bottomrule
  \end{tabular}}\begin{minipage}{\wd0}
  \usebox0\par
  \vspace{4pt}
  \footnotesize \textit{Note.} $\hat{\Delta}_{\text{HL}}$ is the Hodges--Lehmann estimate of the within-sub-sample paired difference between the method and joint-encoding, expressed in the metric's own units. $V$ is the signed-rank test statistic. $p$ is the exact two-sided $p$-value. The favourable column reports the number of sub-samples in which the method outperforms joint-encoding on the row's metric. For \ac{RMSE}, lower values are better, so favourable means $\hat{\Delta}_{\text{HL}} < 0$; for $R^2$ and Spearman $\rho$, higher is better and favourable means $\hat{\Delta}_{\text{HL}} > 0$.
  \end{minipage}
  \end{table}

\begin{figure}[ht]
\centering
\includegraphics[width=\linewidth]{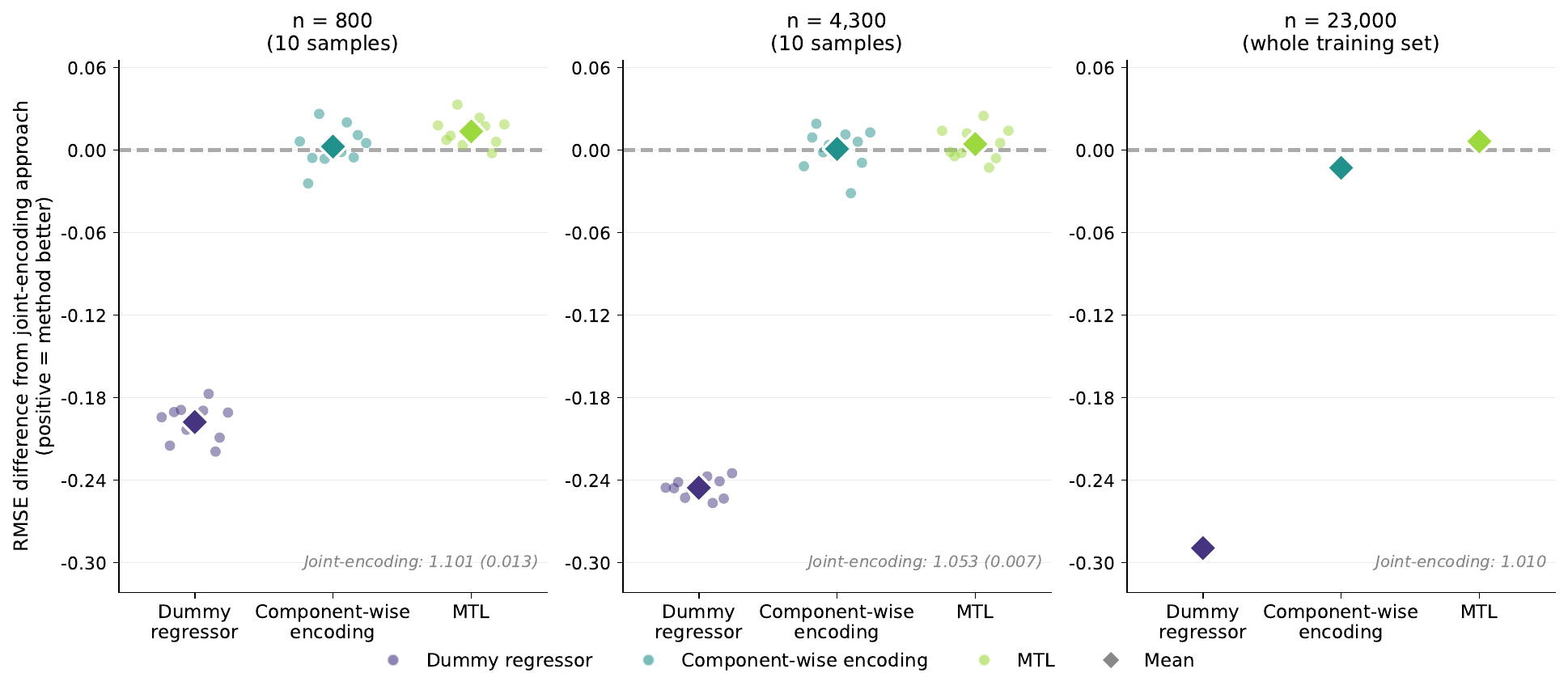}
\caption{Paired differences in \ac{RMSE} between each comparator method (component-wise, \ac{MTL}, and the dummy regressor) and the joint-encoding approach, computed within the same training sub-sample. \rev{All panels share the same y axis; zero is the joint-encoding approach, whose absolute performance is printed in each panel. Diamonds are cell means.}}
\label{fig:paired-rmse}
\end{figure}

\begin{figure}[ht]
\centering
\includegraphics[width=\linewidth]{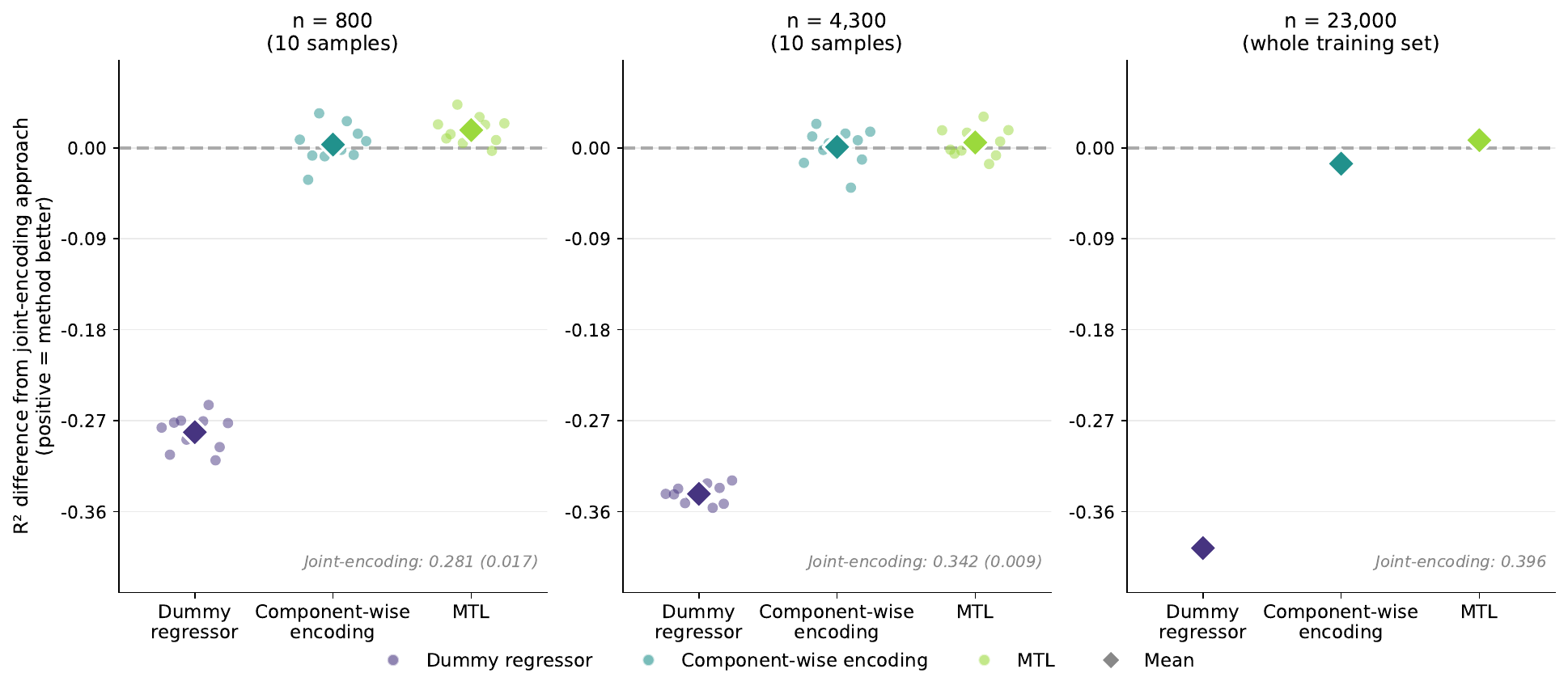}
\caption{Paired differences in $R^2$ between each comparator method (component-wise, \ac{MTL}, and the dummy regressor) and the joint-encoding approach, computed within the same training sub-sample. \rev{All panels share the same y axis; zero is the joint-encoding approach, whose absolute performance is printed in each panel. Diamonds are cell means.}}
\label{fig:paired-r2}
\end{figure}

\begin{figure}[ht]
\centering
\includegraphics[width=\linewidth]{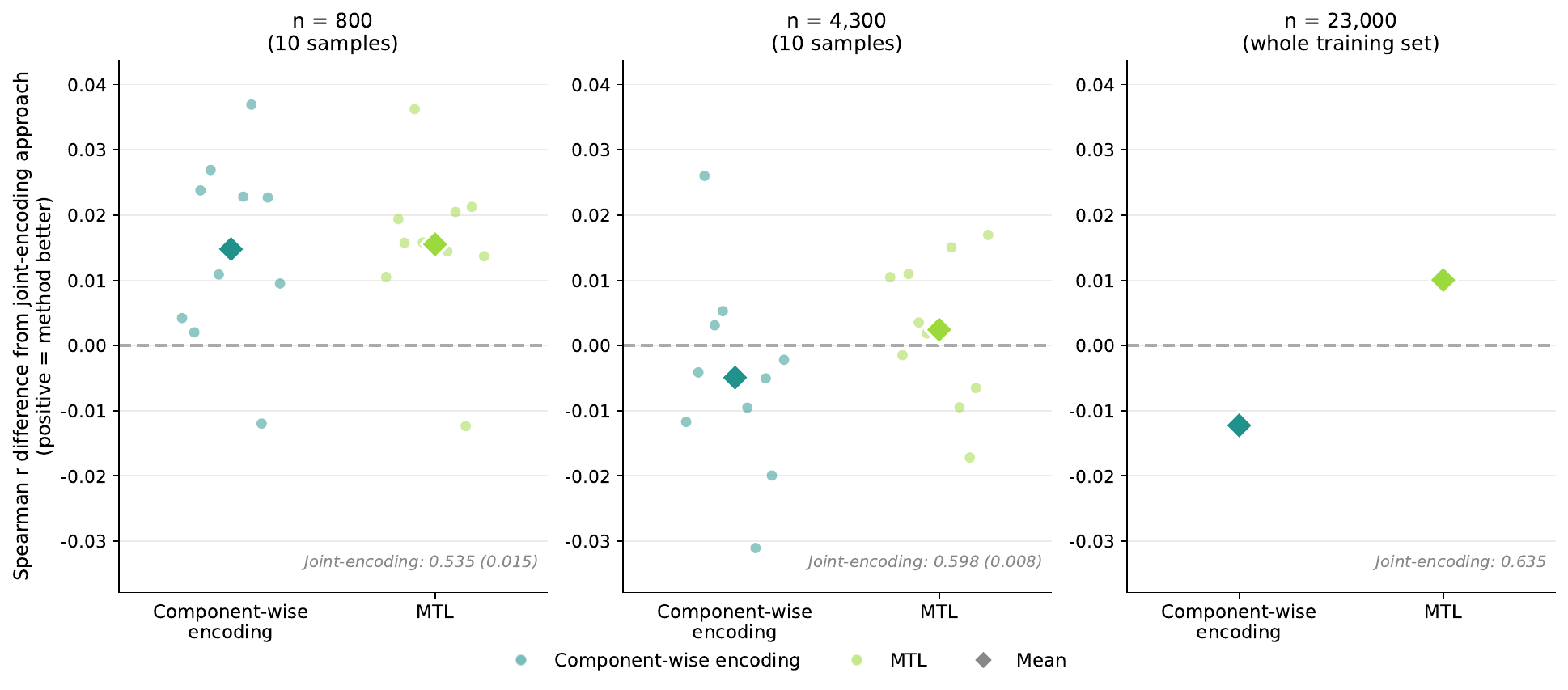}
\caption{Paired differences in Spearman $\rho$ between each comparator method (component-wise and \ac{MTL}) and the joint-encoding approach, computed within the same training sub-sample. The dummy regressor is undefined for correlation since its variance is zero. \rev{All panels share the same y axis; zero is the joint-encoding approach, whose absolute performance is printed in each panel. Diamonds are cell means.}}
\label{fig:paired-spearman}
\end{figure}

Figures~\ref{fig:paired-rmse}--\ref{fig:paired-spearman} visualise the paired differences between a given method and the joint-encoding approach on one of the three reported metrics. \rev{All three panels of a figure share one y axis, whose zero is the joint-encoding approach within the same training sub-sample; the absolute joint-encoding performance that the zero stands for is printed in each panel, so differences can be read both relative to it and across training sizes.}

\section{Discussion}\label{sec:discussion}

This study compared three transformer-based models for response-free item-difficulty prediction within a nested training-size design on a corpus of reading-comprehension \ac{MCQ}s with pseudo-labels anchored to response-pattern-based Rasch difficulty estimates. Three findings emerge.

\rev{First, the transformer-based models presented here are a viable approach to response-free \ac{IDM} for \ac{MCQ}s: every trained method, including the simplest joint-encoding one, improves substantially on the constant dummy reference (the training-set mean of $b$). Clearing that reference is not guaranteed in the field: a}t a training scale comparable to that reported in \citet{yaneva2024}, i.e., 667 items, our methods reduce \ac{RMSE} over the dummy regressor by 14--16\% at $n \approx 800$, against approximately 4\% for the best shared-task team ($0.299$ versus $0.311$ on the test set). The two datasets differ in stimulus domain, item structure, and difficulty scale, so a direct comparison of absolute residuals is not warranted, but a substantially larger relative reduction over the same kind of model-free reference is consistent with our methods capturing more wording-derivable variance. These reductions are obtained without hand-engineered text features or the preprocessing they require.

Second, component-wise encoding shows no detectable benefit over joint encoding under the present design. \rev{The paired differences on \ac{RMSE} and $R^2$ include zero at both training sizes, with only the rank ordering at $n \approx 800$ favouring the component-wise variant, a gain the multi-task variant matches while also improving the other two metrics, so applications that turn on the ordering of items give no reason to prefer component-wise encoding. This holds with a head that is free to combine the component representations non-linearly (Equation~\eqref{eq:structured-head}); with the single linear layer reported in the supplementary material the variant is in fact significantly \emph{worse} than joint-encoding on \ac{RMSE} and $R^2$ at both sizes. We think that the absence of a benefit is therefore not purely an artefact of an under-expressive head. When the components are encoded together, the encoder forms the representation of the passage while it can already ``attend'' to the question and the options, which an encoder ``reading'' each component on its own cannot do \autocite[as alluded to by, e.g.,][]{humeau2020}. What separates joint from component-wise encoding is what the encoder is allowed to see rather than what the head can compute, and no aggregation of the separately pooled vectors recovers information the encoder never had.} We read the findings as evidence that the encoder's self-attention is already recovering the cross-component signal that the component-wise variant aims to enforce manually. \rev{More expressive aggregation than we tried remains possible, although in this literature the shortfall of separately encoded designs is addressed by restoring interaction between the components rather than by enlarging the head that combines them.}

Third, at the smallest training size considered ($n \approx 800$), the multi-task variant produces small but consistent paired improvements over the joint-encoding approach across all three reported metrics (\ac{RMSE}, $R^2$, Spearman $\rho$); this effect diminishes as the training set grows. The improvements are small in absolute terms, but what the results consistently show is that the auxiliary \ac{MCQA} objective provides a training signal that the regression target alone cannot, and that this signal matters precisely in the small-sample regime where end-to-end fine-tuning has the least information to recover from the wording alone or would otherwise learn surface-level shortcuts \autocite{geirhos2020}.

A ceiling on purely text-based prediction necessarily exists, since some difficulty-relevant factors -- test-taking strategies, item-position effects, or visual layout -- leave no trace in the wording. Moreover, the prediction target in our study itself carries two connected layers of noise that bound the achievable accuracy from above. On the PALRACE anchor set, $b_i$ is a Rasch estimate from a finite response sample and is reported with a non-zero standard error; on the broader RACE++ training corpus, $b_i$ is the output of a pseudo-labelling procedure calibrated against PALRACE Rasch \rev{difficulty (correlated with $\rho = .48$}, Section~\ref{sec:datasets}), so an additional component of \ac{LLM} error is stacked on top of the response-based standard error. Any text-based model of $b_i$ is therefore predicting an \emph{estimate}, and both noise sources contribute to the predictive ceiling.

\rev{As an approximate sensitivity calculation, suppose that the model predictions carry no information about response-based Rasch difficulty beyond that conveyed by the pseudo-labels and that the relevant rank associations combine multiplicatively. Under these assumptions, the expected association with Rasch difficulty can be approximated by the product of the model–pseudo-label and pseudo-label–Rasch correlation. Predictions reproducing the pseudo-labels perfectly would therefore correlate .48 with Rasch difficulty. The multi-task correlations of .551 at $n \approx 800$ and .645 at $n \approx 23{,}000$ would then correspond to approximately .26 and .31 with response-based Rasch difficulty. Using the confidence limits $[.42, .53]$ for the pseudo-label agreement gives approximate ranges of $[.23, .29]$ and $[.27, .34]$, respectively. These calculations are illustrative rather than formal confidence intervals, because they do not incorporate all sources of uncertainty and depend on the stated assumptions.
}

\rev{A direct check is available on PALRACE. Fitting the best-performing method, the multi-task variant, to its 787 items with the response-based Rasch estimates as the training target, under ten-fold cross-validation grouped by passage, gives a pooled out-of-fold Spearman correlation of $\rho = .208$ with those estimates. Direct training on several hundred response-based estimates therefore yields accuracy of the same order as the attenuation calculation projects for the pseudo-label route, which supports pseudo-labelling where response data are unavailable.} Rather than a limitation of the approach, these ceilings define the complementary roles that response-free and response-based estimation can play in a calibration workflow: response-free predictions can provide first-pass estimates that response-based information subsequently refines, \rev{though at present levels of accuracy they do not substitute for calibration} \autocite{ulitzsch2025}.

\rev{In the paper, we addressed reading-comprehension \ac{MCQ}s, which sit among the more demanding targets in this literature: for simpler formats such as cloze items, response-free estimates have reached correlation of $.90$ with calibrated difficulties, while for reading comprehension and comparably complex tasks they are markedly less accurate \autocite{ulitzsch2025}. Within that group these items are unusually text-rich, since each is presented with a reading passage rather than as a bare stem, and that passage is continuous English prose, matching the language of the encoder's pre-training corpus. The published evidence settles neither whether the present results extend to items whose difficulty rests on notation, equations or diagrams, nor which item types are intrinsically harder to predict: datasets, difficulty scales and evaluation protocols differ across studies, and the underlying item banks are seldom released, so we believe reported figures cannot be placed on a common footing. The representational challenge such items pose falls on the encoder rather than on the training design, so the methods described here would extend to them with a multimodal encoder in place of the text-only one, once items of that kind can be obtained with calibrated difficulties. That the extension would require only a change of encoder is a property of the end-to-end fine-tuning approach: a pipeline built on hand-engineered text features would need its feature set redesigned for each new item type.}

\rev{Regardless of the representational approach, however, predictive accuracy is not the only criterion for using these models in item development. Predictions should support, rather than mechanically determine, item-development decisions: modifying wording solely to achieve a target predicted difficulty may introduce construct-irrelevant variance unless the affected textual properties are evaluated against a clearly specified construct framework \autocite{MartinkovaPotuznikovaNetik2026_chapter}.}

We recognise several promising directions for future work. A natural next step for the \ac{MCQA} branch, moving beyond the binary correctness target used here, is to replace the auxiliary option classification objective with response-based per-option probabilities, which would let the model receive an explicit signal about distractor competition rather than infer it indirectly from correct-versus-incorrect labels. \rev{As we do not currently have access to a sufficient dataset with per-option probabilities, we examined a model-implied counterpart, in which a separate reading model supplies the option distribution. At the smallest training size, we found no detectable change in difficulty prediction (see the supplementary material for details). A reader model's option probabilities express its own competence rather than the way examinees distribute across the options, so this suggests that the auxiliary task would need a signal grounded in observed responses, without ruling out the approach altogether.} A further possibility would lie in adding an auxiliary head whose objective could be to label distinct wording components (i.e. token-classification task). Next, there remain a number of unexplored techniques beyond design choices implemented in this work (that may have hindered our results) -- two ready alternatives are dynamic loss-balancing schemes \autocite{Cipolla_2018} and gradient-level conflict resolution, such as gradient surgery \autocite{yu2020_gradientsurgery}.

\section{Conclusion}\label{sec:conclusion}

This article compared three transformer fine-tuning architectures for response-free \ac{IDM} on reading-comprehension \ac{MCQ}s under a nested training-size design. Joint encoding of the concatenated item wording is a viable end-to-end alternative to two-stage feature-engineering pipelines; component-wise encoding adds no detectable benefit over joint encoding, consistent with the encoder recovering the \ac{MCQ} component structure through its own self-attention; and the multi-task variant, which adds an auxiliary \ac{MCQA} objective on a shared encoder, yields significant paired improvements at the smallest training size considered ($n \approx 800$).

These results show that transformer fine-tuning enriched with structural or supervisory inductive biases recovers a substantial share of the wording-derivable difficulty signal at training-set sizes that are typically available in applied calibration contexts. Predictions of this form can serve on their own when response data \textit{cannot} be collected, and \rev{where a follow-up is feasible they are best combined with response-based information rather than used in its place}; combining the two sources remains the most robust practice when both are available \autocite[as remarked, e.g., in][]{ulitzsch2025}. A concrete next step is to replace the binary correctness target in the auxiliary head with human response-based per-option probabilities, so that key and distractors competition, which the present formulation can address only indirectly, can shape the encoder directly.

\rev{\section*{Acknowledgments and declarations}

This work was supported by the Grant Agency of Charles University (GAUK), project No.\ 332525, by the Czech Science Foundation project 25-16951S \enquote{Complex analysis of educational measurement data to understand cognitive demands of assessment tasks}, and by the institutional support RVO 67985807.

The authors thank the editors and two anonymous reviewers for their constructive comments on earlier versions of the manuscript.

During the preparation of this manuscript, the authors used Gemini 2.5 Flash as declared in the body of the paper, and Claude Opus 4.7–5 to improve the grammar, syntax, and overall clarity of the text. Additionally, the same tool was utilised to assist with code refactoring, debugging and code review. The authors thoroughly reviewed, verified, and edited all text and code outputs. The authors maintain full accountability for the accuracy, integrity, and originality of the manuscript's intellectual content, and underlying source code.

\subsection*{CRediT authorship contribution statement}

\textbf{JN:} Conceptualization; methodology; software; validation; formal analysis; investigation; data curation; resources; writing -- original draft; writing -- review and editing; visualization; project administration; funding acquisition. \textbf{PM:} Investigation; validation; supervision; writing -- review and editing; funding acquisition.

\section*{Data availability}

The supplementary materials documenting dataset construction, the pseudo-labelling procedure, and the full hyper-parameter specification are available on the Open Science Framework at \url{https://osf.io/pt37f/overview?view_only=48b3615add654e728534e6974483da25}, provided as an anonymous view-only link for the purposes of peer review. Upon acceptance, the same repository will be extended with the analysis code and with the identifiers required to reconstruct the training, validation and test splits from the source corpora. The item wording is not redistributed, since RACE++ and PALRACE are obtained from their original publications; the item identifiers make every record recoverable from those sources.}

\printbibliography

\end{document}